\documentclass[a4paper,twoside]{article}

\usepackage{epsfig}
\usepackage{subcaption}
\usepackage{calc}
\usepackage{amssymb}
\usepackage{amstext}
\usepackage{amsmath}
\usepackage{amsthm}
\usepackage{multicol}
\usepackage{pslatex}
\usepackage{microtype}
\usepackage{multirow}
\usepackage[hidelinks]{hyperref}
\usepackage{enumitem}
\usepackage{natbib}
\usepackage[nolist]{acronym}
\usepackage{SCITEPRESS}     

\begin{document}

\title{Lbl2Vec: An Embedding-Based Approach for Unsupervised Document Retrieval on Predefined Topics}

\author{\authorname{Tim Schopf, Daniel Braun and Florian Matthes}
\affiliation{Department of Informatics, Technical University of Munich, Boltzmannstrasse 3, Garching, Germany}
\email{\{tim.schopf, daniel.braun, matthes\}@tum.de}
}


\abstract{In this paper, we consider the task of retrieving documents with predefined topics from an unlabeled document dataset using an unsupervised approach. The proposed unsupervised approach requires only a small number of keywords describing the respective topics and no labeled document. Existing approaches either heavily relied on a large amount of additionally encoded world knowledge or on term-document frequencies. Contrariwise, we introduce a method that learns jointly embedded document and word vectors solely from the unlabeled document dataset in order to find documents that are semantically similar to the topics described by the keywords. The proposed method requires almost no text preprocessing but is simultaneously effective at retrieving relevant documents with high probability. When successively retrieving documents on different predefined topics from publicly available and commonly used datasets, we achieved an average area under the receiver operating characteristic curve value of 0.95 on one dataset and 0.92 on another. Further, our method can be used for multiclass document classification, without the need to assign labels to the dataset in advance. Compared with an unsupervised classification baseline, we increased F1 scores from 76.6 to 82.7 and from 61.0 to 75.1 on the respective datasets. For easy replication of our approach, we make the developed \texttt{Lbl2Vec} code publicly available as a ready-to-use tool under the 3-Clause BSD license.\footnote{\href{https://github.com/sebischair/Lbl2Vec}{https://github.com/sebischair/Lbl2Vec}}}

\onecolumn \maketitle \normalsize \setcounter{footnote}{0} \vfill

\begin{acronym}
\acro{ke}[KE]{keyword enrichment}
\acro{nlp}[NLP]{Natural Language Processing}
\acro{lsa}[LSA]{latent semantic analysis}
\acro{lda}[LDA]{latent Dirichlet allocation}
\acro{esa}[ESA]{Explicit Semantic Analysis}
\acro{bert}[BERT]{bidirectional encoder representations from transformers}
\acro{dbow}[PV-DBOW]{distributed bag of words version of paragraph vector}
\acro{lof}[LOF]{local outlier factor}
\acro{pcoa}[PCoA]{Principal Coordinates Analysis}
\acro{roc}[ROC]{receiver operating characteristic}
\acro{auc}[AUC]{area under the ROC curve}
\end{acronym}

\section{\uppercase{Introduction}}
\label{sec:introduction}

In this paper, we combine the advantage of an unsupervised approach with the possibility to predefine topics. Precisely, given a large number of unlabeled documents, we would like to retrieve documents related to certain topics that we already know are present in the corpus. This is becoming a common task, considering not only the simplicity of retrieving documents by, e.g., scraping web pages, mails or other sources, but also the labeling cost. For illustration purposes, we imagine the following scenario: we possess a large number of news articles extracted from sports sections of different newspapers and would like to retrieve articles that are related to certain sports, such as hockey, soccer or basketball. Unfortunately, we can only rely on the article texts for this task, as the metadata of the articles contain no information about their content. Initially, this appears like a common text classification task. However, there arise two issues that make the use of conventional classification methods unsuitable. First, we would have to annotate our articles at a high cost, as conventional supervised text classification methods need a large amount of labeled training data \citep{Zhang2020}. Second, we might not be interested in any sports apart from the previously specified ones. However, our dataset of sports articles most likely also includes articles on other sports, such as swimming or running. If we want to apply a supervised classification method, we would either have to annotate even those articles that are of no interest to us or think about suitable previous cleaning steps, to remove unwanted articles from our dataset. Both options would require significant additional expense. 
\newline In this paper, we present the \texttt{Lbl2Vec} approach, which provides the retrieval of documents on predefined topics from a large corpus based on unsupervised learning. This enables us to retrieve the wanted sports articles related to hockey, soccer and basketball only, without having to annotate any data. The proposed \texttt{Lbl2Vec} approach solely relies on semantic similarities between documents and keywords describing a certain topic. Using semantic meanings intuitively matches the approach of a human being and has previously been proven to be capable of categorizing unlabeled texts \citep{Chang2008ImportanceOS}. With this approach, we significantly decrease the cost of annotating data, as we only need a small number of keywords instead of a large number of labeled documents. \newline
\texttt{Lbl2Vec} works by creating jointly embedded word, document, and label vectors. The label vectors are deducted from predefined keywords of each topic. Since label and document vectors are embedded in the same feature space, we can subsequently measure their semantic relationship by calculating their cosine similarity. Based on this semantic similarity, we can decide whether to assign a document to a certain topic or not. \newline
We show that our approach produces reliable results while saving annotation costs and requires almost no text preprocessing steps. To this end, we apply our approach to two publicly available and commonly used document classification datasets. Moreover, we make our \texttt{Lbl2Vec} code publicly available as a ready-to-use tool.

\section{\uppercase{Related Work}}
\label{sec:related_work}

Most related research can be summarized under the notion of \textit{dataless classification}, introduced by \citet{Chang2008ImportanceOS}. Broadly, this includes any approach that aims to classify unlabeled texts based on label descriptions only. Our approach differs slightly from these, as we primarily attempt to retrieve documents on predefined topics from an unlabeled document dataset without the need to consider documents belonging to different topics of no interest. Nevertheless, some similarities, such as the ability of multiclass document classification emerge, allowing a rough comparison of our approach with those from the \textit{dataless classification}, which can further be divided along two dimensions: 1) semi-supervised vs. unsupervised approaches and 2) approaches that use a large amount of additional world knowledge vs. ones that mainly rely on the plain document corpus. \newline
\newline \textbf{Semi-supervised} approaches seek to annotate a small subset of the document corpus unsupervised and subsequently leverage the labeled subset to train a supervised classifier for the rest of the corpus. In one of the earliest approaches that fit into this category, \citet{ko-seo-2000} derive training sentences from manually defined category keywords unsupervised. Then, they used the derived sentences to train a supervised Na\"ive Bayes classifier with minor modifications. Similarly, \citet{liu2004} extracted a subset of documents with keywords and then applied a supervised Na\"ive Bayes-based expectation–maximization algorithm \citep{dempster1977} for classification. \newline
\newline \textbf{Unsupervised} approaches, by contrast, use similarity scores between documents and target categories to classify the entire unlabeled dataset. \citet{haj-yahia-etal-2019-towards} proposed \ac{ke} and subsequent unsupervised classification based on \ac{lsa} \citep{Deerwester1990IndexingBL} vector cosine similarities. Another approach worth mentioning in this context is the pure dataless hierarchical classification used by \citet{Song_Roth_2014} to evaluate different semantic representations. Our approach also fits into this unsupervised dimension, as we do not employ document labels and retrieve documents from the entire corpus based on cosine similarities only. \newline
\newline \textbf{A large amount of additional world knowledge} from different data sources has been widely exploited in many previous approaches to incorporate more context into the semantic relationship between documents and target categories. \citet{Chang2008ImportanceOS} used Wikipedia as source of world knowledge to compute explicit semantic analysis embeddings \citep{gabrilovich2007} of labels and documents. Afterward, they applied the nearest neighbor classification to assign the most likely label to each document. In this regard, their early work had a major impact on further research, which subsequently heavily focused on adding a lot of world knowledge for dataless classification. \citet{yin-etal-2019-benchmarking} used various public entailment datasets to train a \ac{bert} model \citep{devlin-etal-2019-bert} and used the pretrained \ac{bert} entailment model to directly classify texts from different datasets.\newline 
\newline \textbf{Using mainly the plain document corpus} for this task, however, has been rather less researched so far. In one of the earlier approaches, \citet{RaoPK06} derived and assigned document labels based on a k-means word clustering. Besides, \citet{Chen2015DatalessTC} introduce descriptive latent Dirichlet allocation, which could perform classification with only category description words and unlabeled documents, thereby eradicating the need for a large amount of world knowledge from external sources. Since our approach only needs some predefined topic keywords besides the unlabeled document corpus, it also belongs to this category. However, unlike previous approaches that mainly used the plain document corpus, we do not rely on term-document frequency scores but learn new semantic embeddings from scratch, which was inspired by the topic modeling approach of \citet{angelov2020top2vec}.\newline 
\newline
A different related research area addresses \textit{ad-hoc document retrieval}. Approaches related to this area attempt to rank documents based on a relevance score to a specific user query \citep{baeza1999modern}. For instance, \citet{NVSM2018} proposed a neural vector space model that learns document representations unsupervised, and \citet{PV_Ad-hoc_Retrieval2016} introduce a modified paragraph vector model for ad hoc document retrieval. However, our approach differs from these, as we do not want to receive documents based on user queries but topics. Further, we are not particularly interested in ranking within the retrieved documents.

\section{\uppercase{Lbl2Vec Method}}
\label{sec:lbl2vec_method}

\subsection{General Approach}

\begin{figure}[ht]
    \centering
    \includegraphics[width=\linewidth]{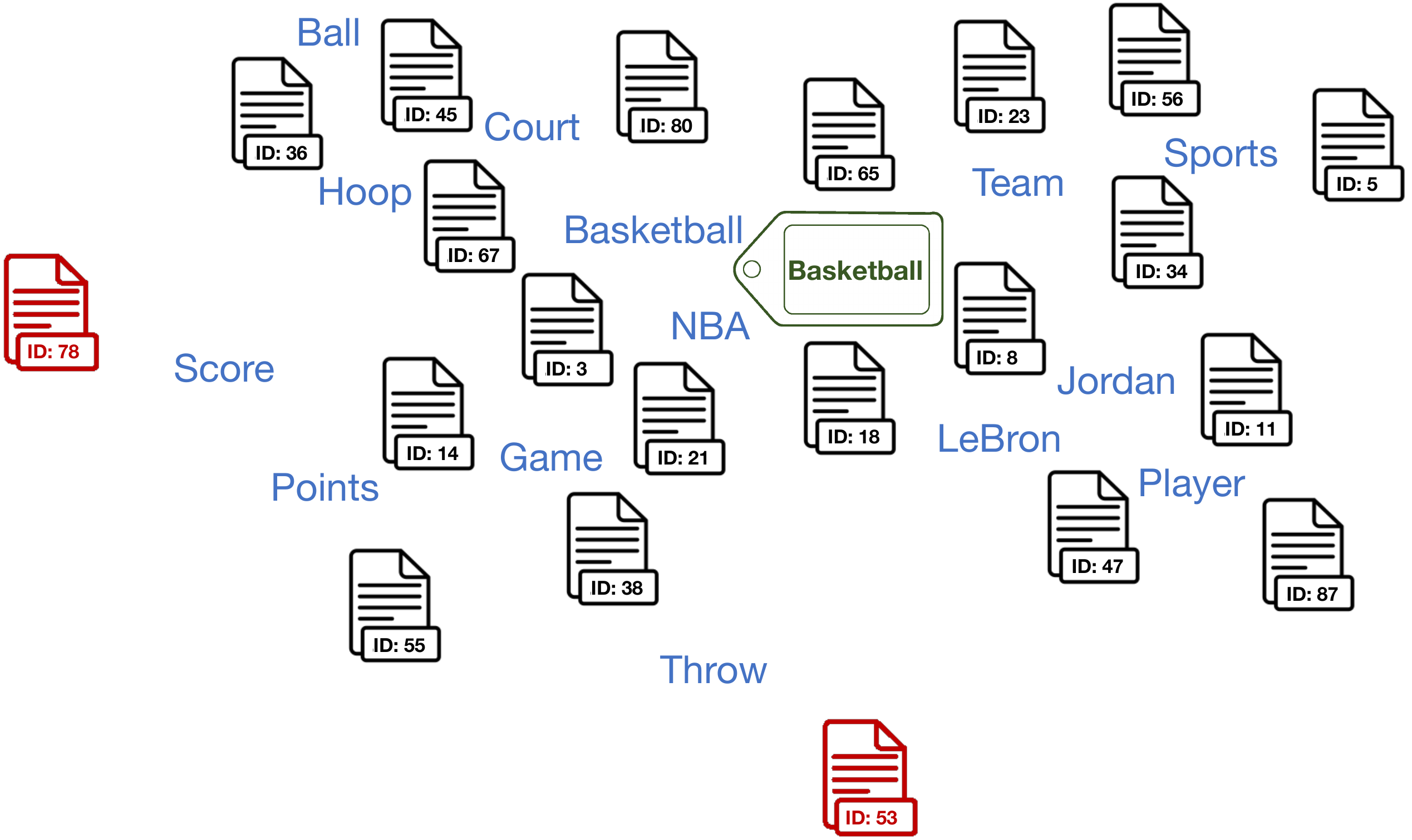}
    \caption{Example illustration of a semantic feature space related to Basketball. Blue: Descriptive keyword embeddings. Black: Document embeddings that are semantically similar to the keywords and each other. Red: Outlier document embeddings. Green: Label embedding.}
    \label{fig:basketball_feature_space}
\end{figure}

In the first step, our \texttt{Lbl2Vec} model learns jointly embedded word vectors $W$ and document vectors $D$ from an unlabeled document corpus. Afterward, we use the embeddings $K \subset W$ of manually defined keywords that describe topics $T$ to learn label embeddings $L$ within the same feature space. Since all learned embeddings $(W,D,L)$ share the same feature space, their distance can be considered their semantic similarity. To learn a label embedding $\vec{l}_i$, we find document embeddings $\vec{d}_{i_1},...,\vec{d}_{i_m}$ that are close to the descriptive keyword embeddings $\vec{k}_{i_1},...,\vec{k}_{i_n}$ of topic $t_i$. Afterward, we compute the centroid of the outlier cleaned document embeddings as the label embedding $\vec{l}_i$ of topic $t_i$. We compute document rather than keyword centroids since our experiments showed that it is more difficult to retrieve documents based on similarities to keywords only, even if they share the same feature space. Moreover, we clean outliers to remove documents that may be related to some of the descriptive keywords but do not properly match the intended topic. As a result, our experiments showed a more accurate label embedding and slightly improved document retrieval performance. Figure \ref{fig:basketball_feature_space} provides an exemplary illustration of the different learned embeddings. After learning, we can consider the distance of label embedding $\vec{l}_{i}$ to an arbitrary document embedding $\vec{d}$ as their semantic similarity. Since we argue that the learned label embeddings are mappings of topics in the semantic feature space, this also represents the semantic similarity between $t_{i}$ and ${d}$. Hence, we use these semantic similarities to finally retrieve those documents related to our predefined topics.

\subsection{Learning Jointly Embedded Semantic Representations}

To train our jointly embedded word and document vectors, we use the paragraph vector framework introduced by \citet{Le-Mikolov-2014}. Since the \ac{dbow} is proven to perform better than its alternative \citep{lau-baldwin-2016-empirical}, we consequently use this architecture. However, \ac{dbow} only trains document embeddings but not word embeddings in its original version. Therefore, we employ a slightly modified implementation that concurrently learns word embeddings and is first mentioned by \citet{dai2015document}. In this modified version, we interleave the \ac{dbow} training with Skip-gram \citep{mikolov2013efficient} word embedding training on the same corpus. As the Skip-gram architecture is very similar to the \ac{dbow} architecture, we simply need to exchange the predicting paragraph vector with a predicting word vector for this purpose. Then, iterative training on the interleaved \ac{dbow} and Skip-gram architectures enable us to simultaneously learn word and document embedding that share the same feature space. \newline
\newline After learning all document and word embeddings, we use the topic keywords for label embedding training. For each topic of interest, we need to manually define at least one keyword that can describe the topic properly. Once all keywords are defined, we perform the following procedure for each topic of interest. By applying
\begin{equation}
    \overline{e}=\frac{1}{n} \sum_{x=1}^{n} \vec{{e}}_{x}
\end{equation}
to calculate a centroid $\overline{e}$ of embeddings ${\vec{{e}}_{1},...,\vec{{e}}_{n}}$, we obtain the centroid $\overline{k}_{i}$ of keyword embeddings for a topic $t_{i}$.
Afterward, we calculate the cosine similarity of $\overline{k}_{i}$ to each $\vec{d} \in D$ and sort the document embeddings in descending order. Beginning at the document embedding with the highest cosine similarity, we now successively add each document embedding to a set of candidate document embeddings $D_{{c}_{i}} \subset D$ that has a high semantic similarity to the descriptive keywords of topic $t_i$. To include only document embeddings with high cosine similarities in $D_{{c}_{i}}$, we additionally need to set values for the three following parameters.
\begin{itemize}[itemsep=2pt,parsep=2pt]
    \item ${s:\{s\in\mathbb{R}|-1\leq s\leq1\}}$ as similarity threshold. Add only document embeddings to $D_{{c}_{i}}$ successively while ${\cos\sphericalangle(\overline{k}_{i},\vec{d})>s}$ is true.
    \item ${d_{min}:\{d_{min}\in\mathbb{N}|1\leq d_{min}\leq d_{max}\leq|D|\}}$ as the minimum number of document embeddings that have to be added to $D_{{c}_{i}}$ successively. This parameter prevents the selection of an insufficient number of documents in case we set $s$ too restrictive.
    \item ${d_{max}:\{d_{max}\in\mathbb{N}|1\leq d_{min}\leq d_{max}\leq|D|\}}$ as the maximum number of document embeddings that may be added to $D_{{c}_{i}}$ successively.
\end{itemize}
To ensure a more accurate label embedding later, we now clean outliers from the resulting set of candidate document embeddings $D_{{c}_{i}}$. Therefore, we apply \ac{lof} \citep{breuning-etal-2000} cleaning. If the \ac{lof} algorithm identifies document embeddings $\vec{d}_{i_{outlier}}$ with significantly lower local density than that of their neighbors, we remove these document embeddings from $D_{{c}_{i}}$. Hence, we receive the set of relevant document embeddings $D_{{r}_{i}} \subseteq D_{{c}_{i}}$ for topic $t_i$. Finally, we compute the centroid of all document embeddings in $D_{{r}_{i}}$ and define this as our label embedding $\vec{l}_i$ of topic $t_i$. Consequently, we obtain jointly embedded semantic representations of words, documents, and topic labels.

\subsection{Receiving Documents on Predefined Topics}

To decide whether the content of document $d$ is semantically similar to a single topic $t_i$, we need to calculate the cosine similarity between document embedding $\vec{d}$ and label embedding $\vec{l}_i$. Subsequently, the affiliation of $d$ to $t_i$ is indicated if $\cos\sphericalangle(\vec{l}_{i},\vec{d})$ exceeds a previously manually defined threshold value ${\alpha_{{t}_{i}}:\{\alpha_{{t}_{i}}\in\mathbb{R}|-1\leq\alpha_{{t}_{i}}\leq1\}}$. Moreover, we can use the cosine similarities for classifying $d$ between multiple different predefined topics ${t_1,...,t_n}$. To achieve this, we assign the label of topic $t_i$ to $d$ if ${\cos\sphericalangle(\vec{l}_{i},\vec{d})=\max(\{\cos\sphericalangle(\vec{l}_{x},\vec{d}):x=1,...,n\})}$. Finally, we can also decide whether a document $d$ does not fit into one of our predefined topics. Therefore, we define threshold values ${\alpha_{{t}_{1}},...,\alpha_{{t}_{n}}}$. In case that $d$ is classified as most similar to topic $t_i$, we discard the label assignment if ${\cos\sphericalangle(\vec{l}_{i},\vec{d})\leq\alpha_{{t}_{i}}}$. As a result, $d$ remains unlabeled, and we assume that the content of this document is unrelated to any of our predefined topics. 

\section{\uppercase{Experiments}}
\label{sec:experiments}

\subsection{Dataset}
We use the two publicly available classification datasets, \textbf{20Newsgroups}\footnote{\href{http://qwone.com/~jason/20Newsgroups}{qwone.com/~jason/20Newsgroups}} and \textbf{AG's Corpus}\footnote{\href{http://groups.di.unipi.it/~gulli/AG_corpus_of_news_articles}{groups.di.unipi.it/{\raise.17ex\hbox{$\scriptstyle\sim$}}gulli/AG\_corpus\_of\_news\_articles}}, described in Table \ref{tab:datasets}. 
\begin{table}[ht]
    \centering
    \scalebox{0.85}{
    \begin{tabular}{|c|c|c|c|}
    \hline
    \textbf{Datasets} & \textbf{\begin{tabular}[c]{@{}c@{}}\#Training\\ documents\end{tabular}} & \textbf{\begin{tabular}[c]{@{}c@{}}\#Test\\ documents\end{tabular}} & \textbf{\#Classes} \\ \hline
    20Newsgroups       & 11314                                                                   & 7532                                                                & 20                 \\ \hline
    AG's Corpus        & 120000                                                                  & 7600                                                                & 4                  \\ \hline
    \end{tabular}}
    \caption{\label{tab:datasets}Summary of the used classification datasets.}
\end{table}
In the following, we consider each class as an independent topic and use the provided class labels solely for evaluation. The 20Newsgroups dataset consists of almost 20,000 documents heterogeneously split across 20 different newsgroup classes. The original AG's Corpus is a collection of over 1 million news articles. We use the version of \citet{zhang-et-al-2015} that construct four evenly distributed classes from the original dataset, resulting in more than 120,000 labeled documents.

\subsection{Keywords Definition}
To determine suitable keywords for each topic represented by a class, we adopt the expert knowledge approach of \citet{haj-yahia-etal-2019-towards}. Hence, we emulate human experts ourselves, that define some initial keywords based on the class descriptions only. Then, we randomly select some documents from each class to further derive some salient keywords. In the case of a strict unsupervised setting with completely unlabeled datasets, human experts might describe a topic with keywords based on their specific domain knowledge alone and without necessarily being familiar with the document contents. 

\subsection{Model Training}
For model training, we need to convert all document words and topic keywords to lowercase. To finish our short preprocessing, we only have to tokenize the documents and assign IDs to them. For each dataset, we train an individual model. Accordingly, we pass the corresponding preprocessed documents and defined keywords to its own model. For our models to learn suitable embeddings, we need to set the hyperparameter values prior to training. Therefore, we conduct a short manual hyperparameter optimization by training \texttt{Lbl2Vec} models on the respective training datasets and evaluating the performance on the test datasets, which allows us to learn more precise embeddings while simultaneously avoiding overfitting. In the case of completely unlabeled datasets, the given standard hyperparameters can be used. The only significant hyperparameter setting difference between the two models, resulting from our hyperparameter optimization, is that we set a similarity threshold of ${s=0.30}$ and ${s=0.43}$ for the AG's Corpus and 20Newsgroups models, respectively. For both models, we choose ${d_{min}=100}$, ${d_{max}=|D|}$, and 10 as the number of epochs for \ac{dbow} training. As we use an unsupervised approach, we train our final models, similar to \citet{haj-yahia-etal-2019-towards}, on the entire corpora of the respective aggregated training and test datasets.

\subsection{Topic Representation Analysis}
\begin{figure}[ht]
    \centering
    \includegraphics[width=\linewidth]{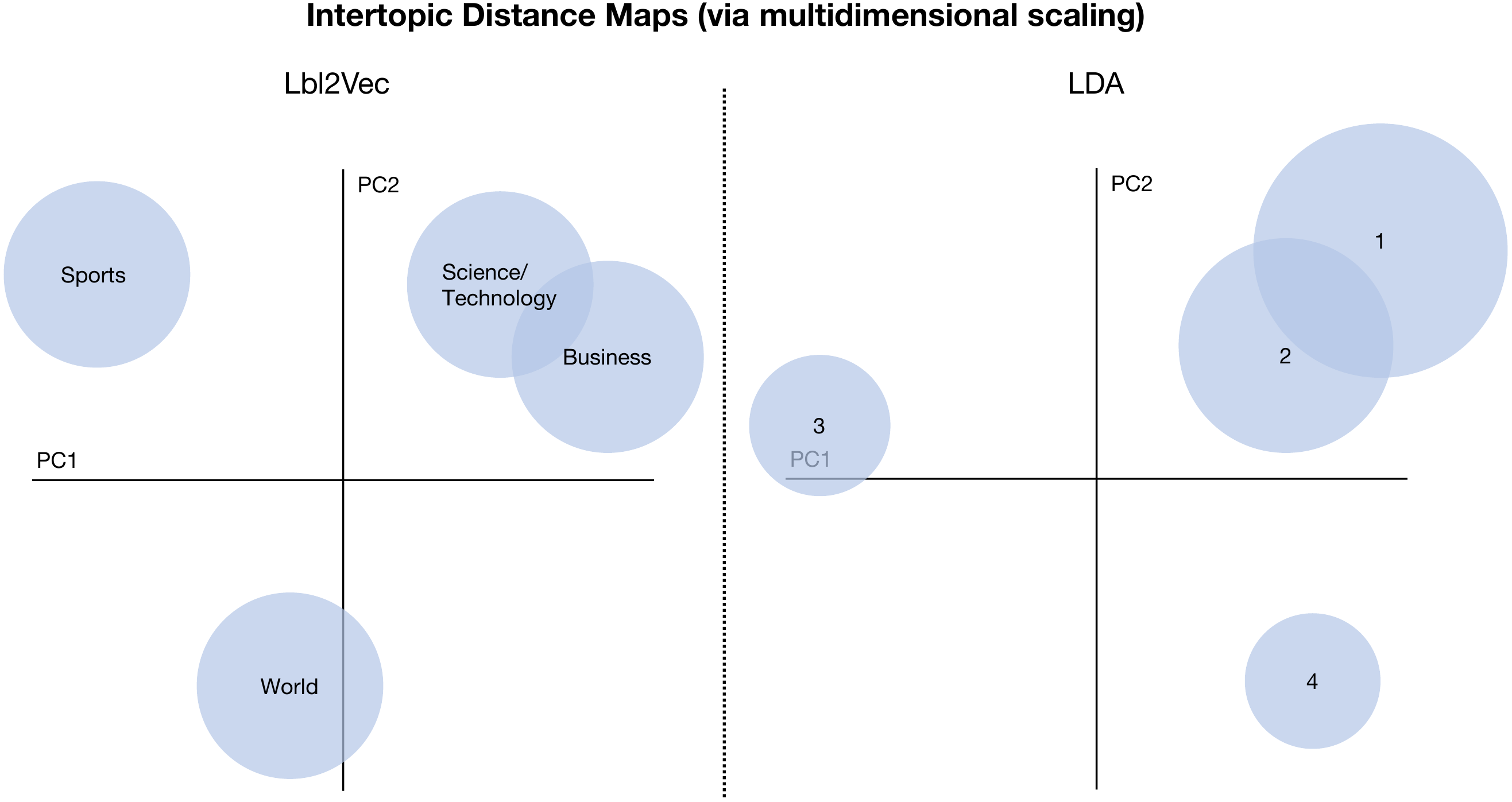}
    \caption{Visualization of Lbl2Vec and LDA topic representation capabilities based on AG's Corpus. Each circle represents a topic, whereas each topic, in turn, consists of several documents classified as related by the respective models. The size of the circles is proportional to the relative occurrence of the respective topic in the corpus. Distances between circles represent semantic inter-topic similarities.}
    \label{fig:pyLDAvis_visualization}
\end{figure}

We want to evaluate whether our \texttt{Lbl2Vec} approach is capable of adequately modeling predefined topics and thereby can return documents related to them. For that, we classify all documents in the AG's Corpus using our pretrained \texttt{Lbl2Vec} model. Afterward, we define the documents assigned to the same class by our model as one topic and analyze these topics using LDAvis \citep{sievert-shirley-2014-ldavis}. In addition, we compare the modeling capabilities on predefined topics of our \texttt{Lbl2Vec} approach to a common topic modeling approach. To this end, we apply \ac{lda} \citep{lda} with $K=4$ number of topics to the same dataset and visualize the modeled topics. Figure \ref{fig:pyLDAvis_visualization} shows that the \ac{lda} model finds two similar and two dissimilar topics.
However, the topic sizes are distributed very heterogeneously, which contrasts with the uniform distribution of documents across all classes in the AG's Corpus. As opposed to this, our \texttt{Lbl2Vec} model finds topics that are equally sized, which is aligned with the underlying AG's Corpus. Further, the topics \textit{Science/Technology} and \textit{Business} are similar, whereas \textit{Sports} and \textit{World} are highly dissimilar to all other topics.
\begin{table}[ht]
    \centering
    \resizebox{\linewidth}{!}{%
    \begin{tabular}{|c|l|}
    \hline
    \multicolumn{2}{|c|}{\textbf{LDA}}                                                                                                             \\ \hline
    Topic 1 & \begin{tabular}[c]{@{}l@{}}oil; crude; prices; microsoft; windows; \\ dollar; reuters; barrel; stocks; yukos;\end{tabular}            \\ \hline
    Topic 2 & \begin{tabular}[c]{@{}l@{}}ccia; thunderbird; generali; macau; cheetham;\\ backman; hauritz; pizarro; rituxan; abdicate;\end{tabular} \\ \hline
    Topic 3 & \begin{tabular}[c]{@{}l@{}}orton; mashburn; bender; kwame; pippen;\\ attanasio; elliss; icelandair; lefors; stottlemyre;\end{tabular} \\ \hline
    Topic 4 & \begin{tabular}[c]{@{}l@{}}wiltord; perrigo; quetta; dione; mattick;\\ olympiad; panis; agis; bago; cracknell;\end{tabular}           \\ \hline
    \end{tabular}%
    }\caption{\label{tab:lda_words}Top 10 most relevant terms for each topic of the \ac{lda} model; we use the LDAvis relevance with $\lambda=0.1$.}
\end{table}
Table \ref{tab:lda_words} indicates that a standard topic modeling approach like \ac{lda} cannot model predefined topics such as the AG's Corpus classes. The most relevant terms of the \ac{lda} topics mainly consist of different entities and do not allow us to relate the modeled topics to the AG's Corpus classes. 
\begin{table}[ht]
\centering
    \resizebox{\linewidth}{!}{%
    \begin{tabular}{|c|l|}
    \hline
    \multicolumn{2}{|c|}{\textbf{Lbl2Vec}}                                                                                                        \\ \hline
    World    & \begin{tabular}[c]{@{}l@{}}iraq; killed; minister; prime; military;\\ palestinian; minister; israeli; troops; darfur;\end{tabular} \\ \hline
    Sports   & \begin{tabular}[c]{@{}l@{}}cup; coach; sox; league; championship;\\ yankees; champions; win; season; scored;\end{tabular}          \\ \hline
    Business & \begin{tabular}[c]{@{}l@{}}stocks; fullquote; profit; prices; aspx;\\ quickinfo; shares; earnings; investor; oil;\end{tabular}     \\ \hline
    \begin{tabular}[c]{@{}c@{}}Science/\\ Technology\end{tabular} &
      \begin{tabular}[c]{@{}l@{}}microsoft; windows; users; desktop; music;\\ linux; version; apple; search; browser;\end{tabular} \\ \hline
    \end{tabular}%
    }\caption{\label{tab:lbl2vec_words}Top 10 most relevant terms for each topic of the \texttt{Lbl2Vec} model; we use the LDAvis relevance with $\lambda=0.1$.}
\end{table}
However, from Table \ref{tab:lbl2vec_words} we can conclude that our \texttt{Lbl2Vec} model can capture the semantic meaning of each predefined topic very well. In addition, the occurrence of technology companies such as Microsoft and Apple in the \textit{Science/Technology} topic explains the similarity to the \textit{Business} topic, as such companies are also highly relevant in a business context. 

\subsection{Multiclass Document Classification Results}
When using our trained models to classify the entire document corpus of each dataset, we achieve the results stated in Table \ref{tab:classification_results}. We compared our models with a recent fully unsupervised text classification approach and a supervised baseline classifier.
\begin{table}[ht]
\centering
\resizebox{\linewidth}{!}{%
    \begin{tabular}{lllllll}
    \multicolumn{1}{c}{\multirow{2}{*}{\textbf{Method}}} & \multicolumn{3}{l}{\textbf{AG's Corpus}} & \multicolumn{3}{l}{\textbf{20Newsgroups}} \\ \cline{2-7} 
    \multicolumn{1}{c}{}     & F1   & Prec. & Rec. & F1   & Prec. & Rec. \\ \hline
    \ac{ke} + \ac{lsa} & 76.6 & 76.8  & 76.6 & 61.0 & 71.1  & 57.8 \\
    Lbl2Vec                  & 82.7 & 82.7  & 82.7 & 75.1 & 75.1  & 75.1 \\ \hline
    Supervised Na\"ive Bayes   & 89.8 & 89.8  & 89.9 & 85.0 & 87.1  & 85.4
    \end{tabular}%
    }\caption{\label{tab:classification_results}Performance of our \texttt{Lbl2Vec} models when classifying all documents in the respective corpus. \ac{ke} + \ac{lsa} refers to the best possible fully unsupervised classification results of \citet{haj-yahia-etal-2019-towards} on the datasets. The last row states their baseline classification results of a supervised multinomial Na\"ive Bayes approach. As we used micro-averaging to calculate our classification metrics, we realized equal F1, Precision, and Recall scores within each model.}
\end{table}
First, we observed that our \texttt{Lbl2Vec} models significantly outperformed the recent \ac{ke} + \ac{lsa} approach for each metric. This success indicated that using our jointly created embeddings for unsupervised classification is more suitable than using term-document frequencies on which \ac{lsa} is heavily reliant. Moreover, the results showed that our \texttt{Lbl2Vec} approach allowed for unsupervised classification in case the labeling effort was estimated to be more expensive than the benefit of a more accurate classification. However, comparing our approach to the supervised baseline results, we observed that providing labels for each document is paramount if highly accurate classification results are required. 

\subsection{Document Retrieval Evaluation}\label{ROC_eval_section}
\begin{figure}[ht]
    \centering
    \includegraphics[width=\linewidth]{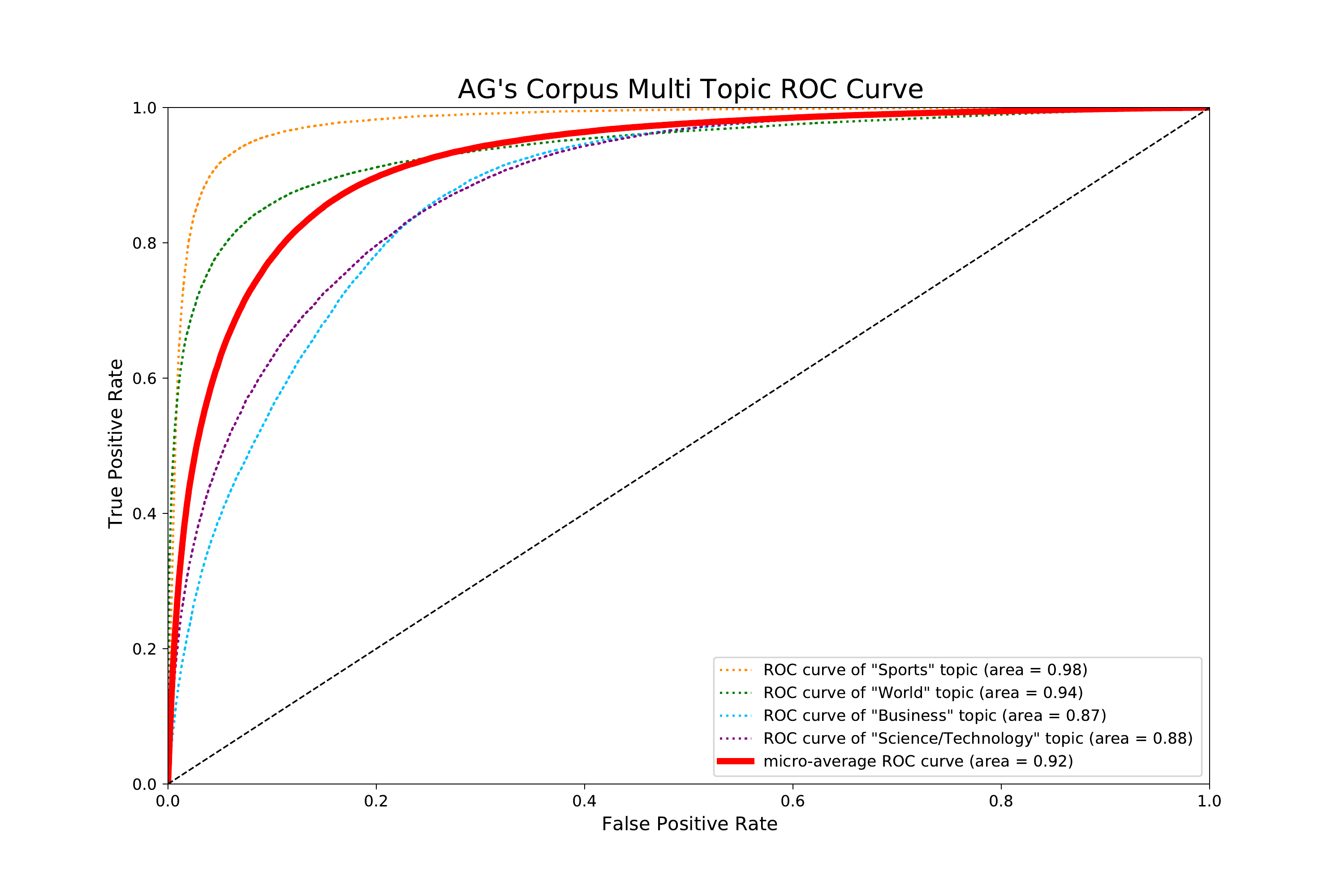}
    \caption{\ac{roc} curves of the \texttt{Lbl2Vec} model trained on the AG's Corpus.}
    \label{fig:AGCorpus_ROC}
\end{figure}

\begin{figure}[ht]
    \centering
    \includegraphics[width=\linewidth]{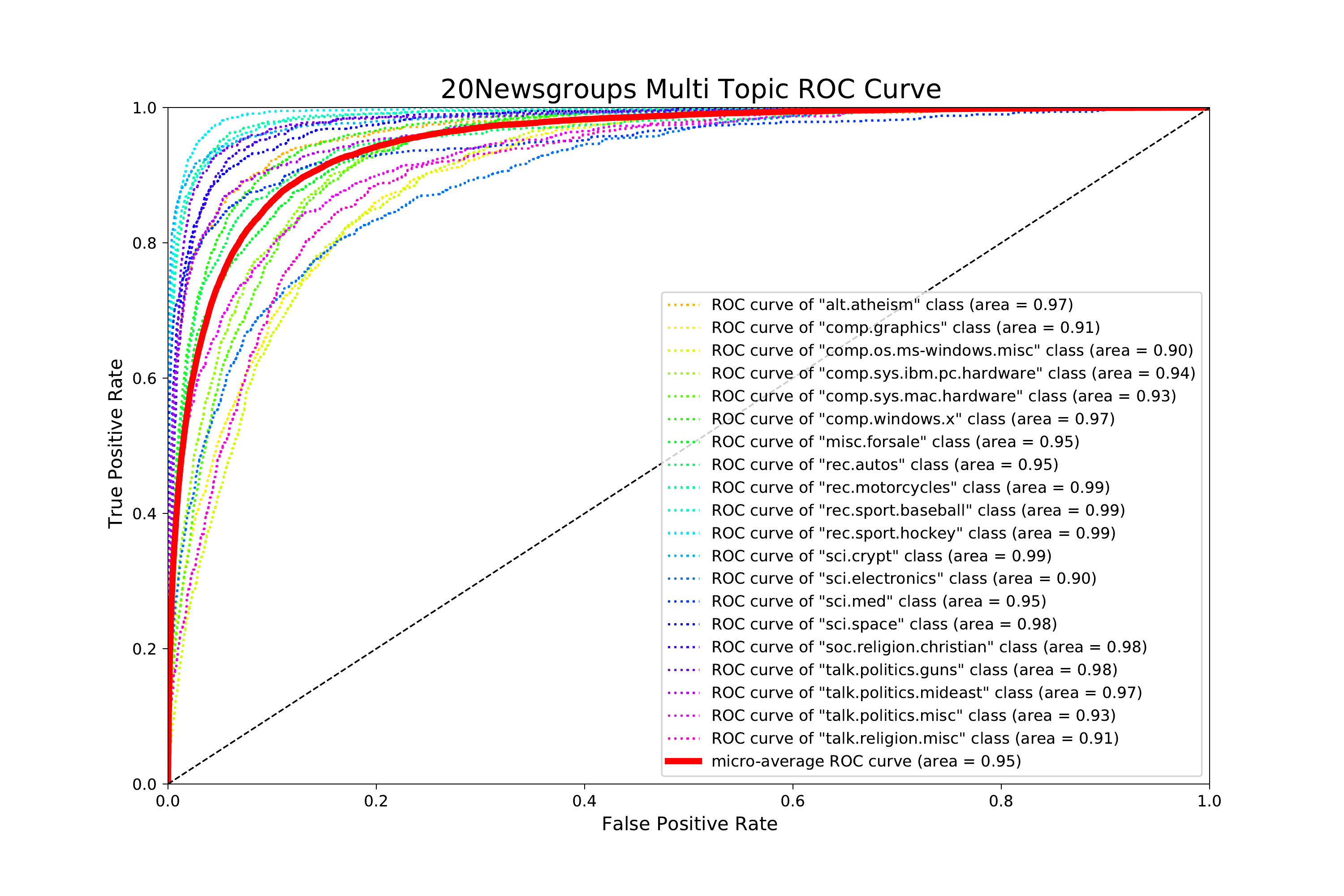}
    \caption{\ac{roc} curves of the \texttt{Lbl2Vec} model trained on the 20Newsgroups.}
    \label{fig:20Newsgroups_ROC}
\end{figure}

One of the main features of our \texttt{Lbl2Vec} approach is retrieving related documents on a single or multiple predefined topics without actually having to consider any further topics contained in the dataset that may not be of interest. For both datasets, we see each class as an independent topic. Therefore, we can use our trained \texttt{Lbl2Vec} models to retrieve topic-related documents for each class independently. When adjusting the topic similarity thresholds ${\alpha_{t_{1}},...,\alpha_{t_{m}}}$ for each topic ${t_1,...,t_m}$ in the respective datasets, we can observe the \ac{roc} curves in Figures \ref{fig:AGCorpus_ROC} and \ref{fig:20Newsgroups_ROC}.
By adjusting the topic similarity parameter $\alpha$ to be closer to 1, we can reduce the false positive rate and retrieve proportionally more documents that are truly related to a topic. Figure \ref{fig:AGCorpus_ROC} shows that the topics, \textit{Business} and \textit{Science/Technology}, have the lowest \ac{auc} values of all topics within the AG's Corpus. Further, we know from Figure \ref{fig:pyLDAvis_visualization} that these topics are similar. Hence, we infer that it is hard for our \texttt{Lbl2Vec} approach to distinguish between related topics. However, the better \ac{auc} values for the \textit{Sports} and \textit{World} topics in Figure \ref{fig:AGCorpus_ROC} and their distance to other topics in Figure \ref{fig:pyLDAvis_visualization} show that our \texttt{Lbl2Vec} approach can create suitable topic representations given the absence of other similar topics in the dataset. The micro-average \ac{roc} curves of Figures \ref{fig:AGCorpus_ROC} and \ref{fig:20Newsgroups_ROC} indicate that, if we want to achieve a false positive rate of less than 1\% on average, we retrieve $\approx20\%$ of documents that are truly relevant for a topic. Therefore, we argue that our \texttt{Lbl2Vec} approach can sample a small dataset with high precision from a large corpus of documents. This smaller dataset can then be used, for example, as a starting point for a subsequent semi-supervised classification approach.

\subsection{Keywords Analysis}

We are additionally interested in how the choice of keywords affects our \texttt{Lbl2Vec} results. Since the keywords also directly affect the predefined topics, this simultaneously involves the analysis of topic distributions. We conduct some hypothesis tests to address the question of what characterizes good keywords and topics. For all our tests, we use the defined keywords of each topic from the concatenation of the two datasets to compute correlation coefficients and determine a significance level of 0.05. We choose Kendall's $\tau$ as our correlation coefficient to measure monotonic relationships. It is robust against outliers and small datasets.\newline
\newline First, we test whether the trained \texttt{Lbl2Vec} model is subsequently better able to distinguish topic-related documents from unrelated ones the more topic-related keywords are used to describe a topic. This test assumes that more accurate descriptions of topics also require more topic-related keywords. Accordingly, we define our null hypothesis $H_{0}^{(1)}$ as the \ac{auc} values of topics modeled by \texttt{Lbl2Vec} are unrelated to the number of topic-related predefined keywords and our alternative hypothesis $H_{a}^{(1)}$ as the \ac{auc} values of topics modeled by \texttt{Lbl2Vec} are positively related to the number of topic-related predefined keywords. 
\begin{table}[ht]
\centering
\resizebox{0.65\linewidth}{!}{%
    \begin{tabular}{|l|l|}
    \hline
    \textbf{Correlation coefficient} & \textbf{p-value} \\ \hline
    Kendall's $\tau$ = 0.19             & 0.20             \\ \hline
    \end{tabular}%
    }\caption{\label{tab:num_keywords_corr} Correlation values that measure the relationship between ${X_{1}=\textrm{number of defined topic keywords}}$ and ${Y=\textrm{\ac{auc}}}$ value of a topic. $X_{{1}_{min}}=10$ and $X_{{1}_{max}}=44$.}
\end{table}
At first glance, the correlation coefficient in Table \ref{tab:num_keywords_corr} suggested a tendency toward a slightly positive correlation. However, the p-value exceeded our defined significance level. Therefore, our test results were statistically insignificant, hence we cannot reject $H_{0}^{(1)}$. Consequently, we found no support for the assumption that \texttt{Lbl2Vec} can yield better topic models if we use more topic-related keywords, as there is insufficient evidence to infer a relationship between $X_1$ and $Y$.\newline
\newline Second, we asses whether using many similar keywords to describe a topic provides a better distinction from other topics than using many dissimilar keywords. As a result, we anticipate \texttt{Lbl2Vec} topic models are better at distinguishing topic-related documents from unrelated ones if we define mostly similar keywords for a single topic. To test this, we initially define the average intratopic similarity of keyword embeddings $K_{i}$ of a topic $t_{i}$ as follows:
\begin{equation}
    \Delta(i)=\frac{\sum\limits_{\substack{\vec{k}_{{i}_x},\vec{k}_{{i}_y} \in K_{i}\\\vec{k}_{{i}_x}\neq\vec{k}_{i_{y}}}}\cos\sphericalangle(\vec{k}_{{i}_x},\vec{k}_{{i}_y})}{|K_{i}|\cdot(|K_{i}|-1)}
\end{equation}
Subsequently, we determine our null hypothesis $H_{0}^{(2)}$ as the \ac{auc} values of topics modeled by \texttt{Lbl2Vec} are unrelated to the average intratopic similarity of topic keywords and our alternative hypothesis $H_{a}^{(2)}$ as the \ac{auc} values of topics modeled by \texttt{Lbl2Vec} are positively related to the average intratopic similarity of topic keywords. 
\begin{table}[ht]
\centering
\resizebox{0.65\linewidth}{!}{%
    \begin{tabular}{|l|l|}
    \hline
    \textbf{Correlation coefficient} & \textbf{p-value} \\ \hline
    Kendall's $\tau$ = 0.33             & 0.02             \\ \hline
    \end{tabular}%
    }\caption{\label{tab:intratopic_corr} Correlation values that measure the relationship between ${X_2=}$ average intratopic similarity of topic keywords and ${Y=\textrm{\ac{auc}}}$ value of a topic. $X_{2_{min}}=0.15$ and $X_{2_{max}}=0.37$.}
\end{table}
Based on the p-value in Table \ref{tab:intratopic_corr}, we rejected $H_{0}^{(2)}$ and from the correlation coefficient, we concluded a statistically significant medium positive correlation between $X_2$ and $Y$. From this evidence, we found support for our original assumption that using similar keywords to describe a topic yields better \texttt{Lbl2Vec} models. \newline
\newline The third test is based on our observation from Subsection \ref{ROC_eval_section}, that \texttt{Lbl2Vec} models more accurate representations of topics dissimilar to all other topics within a dataset. We further investigate this aspect, by examining whether topic keywords highly dissimilar to all other topic keywords allow \texttt{Lbl2Vec} to model more precise topic representations. For this test, we define the average intertopic similarity of keyword embeddings $K_i$ of a topic $t_i$ as 
\begin{equation}
    \delta(i)=\frac{1}{(|T|-1)}\sum_{\substack{n\neq i}}^{(|T|-1)} \frac{\sum\limits_{\substack{\vec{k}_{{i}_x}\in K_{i} \\ \vec{k}_{n_{y}}\in K_{n}}}\cos\sphericalangle(\vec{k}_{{i}_x},\vec{k}_{{n}_y})}{|K_{i}|\cdot|K_{n}|}.
\end{equation}
Afterward, we define our null hypothesis $H_{0}^{(3)}$ as the \ac{auc} values of topics modeled by \texttt{Lbl2Vec} are unrelated to the average intertopic similarity of topic keywords and our alternative hypothesis $H_{a}^{(3)}$ as the \ac{auc} values of topics modeled by \texttt{Lbl2Vec} are negatively related to the average intertopic similarity of topic keywords. From Table \ref{tab:intertopic_corr}, we concluded a moderate negative monotonic relationship between ${X_3}$ and $Y$.
\begin{table}[ht]
\centering
\resizebox{0.65\linewidth}{!}{%
    \begin{tabular}{|l|l|}
    \hline
    \textbf{Correlation coefficient} & \textbf{p-value} \\ \hline
    Kendall's $\tau$ = -0.35             & 0.02             \\ \hline
    \end{tabular}%
    }\caption{\label{tab:intertopic_corr} Correlation values that measure the relationship between ${X_3=}$ average intertopic similarity of topic keywords and ${Y=\textrm{\ac{auc}}}$ value of a topic. $X_{3_{min}}=0.07$ and $X_{3_{max}}=0.11$.}
\end{table}
Moreover, from the p-value, we infer that our third hypothesis test is statistically significant and we can reject $H_{0}^{(3)}$. The defined topic keywords provide the foundation for the subsequent \texttt{Lbl2Vec} feature space embedding of a topic. The feature space location, in turn, determines the similarity of topics to each other. Accordingly, the dissimilarity of topic keywords transfers to the resulting \texttt{Lbl2Vec} topic representations and vice versa. Hence, in this statistically significant inter-topic keywords similarity test, we found further support for our earlier observation that topics dissimilar to all other topics may be modeled more precisely by \texttt{Lbl2Vec}. Consequently, to obtain a more precise topic representation by \texttt{Lbl2Vec}, we need to define topic keywords making them as dissimilar as possible to the keywords of other topics. \newline

\section{\uppercase{Conclusion}}
\label{sec:conclusion}

In this work, we introduced \texttt{Lbl2Vec}, an approach to retrieve documents from predefined topics unsupervised. It is based on jointly embedded word, document, and label vectors learned solely from an unlabeled document corpus. We showed that \texttt{Lbl2Vec} yields better fitting models of predefined topics than conventional topic modeling approaches, such as \ac{lda}. Further, we demonstrated that \texttt{Lbl2Vec} allowed for unsupervised document classification and could retrieve documents on predefined topics with high precision by adjusting the topic similarity parameter $\alpha$. Finally, we analyzed how to define keywords that yield good \texttt{Lbl2Vec} models and concluded that we need to aim for high intratopic similarities and high intertopic dissimilarities of keywords. \texttt{Lbl2Vec} facilitates the retrieval of documents on predefined topics from an unlabeled document corpus, avoiding costly labeling work. We made our \texttt{Lbl2Vec} code as well as the data publicly available.

\section{\uppercase{Ethical Considerations}}
\label{sec:ethical_considerations}

We provide our work in good faith and in accordance with the ACL Code of Ethics\footnote{\href{https://www.aclweb.org/portal/content/acl-code-ethics}{https://www.aclweb.org/portal/content/acl-code-ethics}}. However, our approach depends heavily on the underlying data. Therefore, users should preprocess the targeted datasets according to the ethics' guidelines to prevent discrimination in the modeled topics. Further, our approach is heavily prone to bias introduced by the human expert defining the keywords and unprotected against intentional misuse, allowing malicious users to abuse the retrieved topics. Another concern, as with many models, is the environmental and financial costs incurred in the training process. Although such costs are naturally involved in our case, they are quite low compared with current state-of-the-art language models. Thus, our approach is comparably environmentally friendly and enables financially disadvantaged users to conduct further research.

\section*{ACKNOWLEDGEMENTS}
\label{sec:acknowledgments}

The authors would like to thank Thomas Kinkeldei of ROKIN for his contributions to this paper. \newline
This work has been supported by funds from the Bavarian Ministry of Economic Affairs, Regional Development and Energy as part of the program “Bayerischen Verbundförderprogramms (BayVFP) – Förderlinie Digitalisierung – Förderbereich Informations- und Kommunikationstechnik”.

\bibliographystyle{apalike}
{\small
\bibliography{anthology,custom}}

\begin{thebibliography}{}

\bibitem[Ai et~al., 2016]{PV_Ad-hoc_Retrieval2016}
Ai, Q., Yang, L., Guo, J., and Croft, W.~B. (2016).
\newblock Improving language estimation with the paragraph vector model for
  ad-hoc retrieval.
\newblock In {\em Proceedings of the 39th International ACM SIGIR Conference on
  Research and Development in Information Retrieval}, SIGIR '16, page
  869–872, New York, NY, USA. Association for Computing Machinery.

\bibitem[Angelov, 2020]{angelov2020top2vec}
Angelov, D. (2020).
\newblock Top2vec: Distributed representations of topics.

\bibitem[Baeza-Yates and Ribeiro-Neto, 1999]{baeza1999modern}
Baeza-Yates, R. and Ribeiro-Neto, B. (1999).
\newblock {\em Modern information retrieval}, volume 463.
\newblock ACM press New York.

\bibitem[Blei et~al., 2003]{lda}
Blei, D., Ng, A., and Jordan, M. (2003).
\newblock Latent dirichlet allocation.
\newblock {\em Journal of Machine Learning Research}, 3:993--1022.

\bibitem[Breunig et~al., 2000]{breuning-etal-2000}
Breunig, M.~M., Kriegel, H.-P., Ng, R.~T., and Sander, J. (2000).
\newblock Lof: Identifying density-based local outliers.
\newblock In {\em Proceedings of the 2000 ACM SIGMOD international conference
  on Management of data}, page 93–104, New York, NY, USA. Association for
  Computing Machinery.

\bibitem[Chang et~al., 2008]{Chang2008ImportanceOS}
Chang, M.-W., Ratinov, L.-A., Roth, D., and Srikumar, V. (2008).
\newblock Importance of semantic representation: Dataless classification.
\newblock In {\em Proceedings of the Twenty-Third AAAI Conference on Artificial
  Intelligence}, pages 830--835.

\bibitem[Chen et~al., 2015]{Chen2015DatalessTC}
Chen, X., Xia, Y., Jin, P., and Carroll, J. (2015).
\newblock Dataless text classification with descriptive lda.
\newblock In {\em Proceedings of the Twenty-Ninth AAAI Conference on Artificial
  Intelligence}.

\bibitem[Dai et~al., 2015]{dai2015document}
Dai, A.~M., Olah, C., and Le, Q.~V. (2015).
\newblock Document embedding with paragraph vectors.

\bibitem[Deerwester et~al., 1990]{Deerwester1990IndexingBL}
Deerwester, S., Dumais, S.~T., Furnas, G.~W., Landauer, T.~K., and Harshman, R.
  (1990).
\newblock Indexing by latent semantic analysis.
\newblock {\em Journal of the American Society for Information Science},
  41(6):391--407.

\bibitem[Dempster et~al., 1977]{dempster1977}
Dempster, A.~P., Laird, N.~M., and Rubin, D.~B. (1977).
\newblock Maximum likelihood from incomplete data via the em algorithm.
\newblock {\em Journal of the Royal Statistical Society: Series B
  (Methodological)}, 39(1):1--22.

\bibitem[Devlin et~al., 2019]{devlin-etal-2019-bert}
Devlin, J., Chang, M.-W., Lee, K., and Toutanova, K. (2019).
\newblock {BERT}: Pre-training of deep bidirectional transformers for language
  understanding.
\newblock In {\em Proceedings of the 2019 Conference of the North {A}merican
  Chapter of the Association for Computational Linguistics: Human Language
  Technologies, Volume 1 (Long and Short Papers)}, pages 4171--4186,
  Minneapolis, Minnesota. Association for Computational Linguistics.

\bibitem[Gabrilovich and Markovitch, 2007]{gabrilovich2007}
Gabrilovich, E. and Markovitch, S. (2007).
\newblock Computing semantic relatedness using wikipedia-based explicit
  semantic analysis.
\newblock In {\em Proceedings of the 20th International Joint Conference on
  Artifical Intelligence}, IJCAI'07, page 1606–1611, San Francisco, CA, USA.
  Morgan Kaufmann Publishers Inc.

\bibitem[Gysel et~al., 2018]{NVSM2018}
Gysel, C.~V., de~Rijke, M., and Kanoulas, E. (2018).
\newblock Neural vector spaces for unsupervised information retrieval.
\newblock {\em ACM Trans. Inf. Syst.}, 36(4).

\bibitem[Haj-Yahia et~al., 2019]{haj-yahia-etal-2019-towards}
Haj-Yahia, Z., Sieg, A., and Deleris, L.~A. (2019).
\newblock Towards unsupervised text classification leveraging experts and word
  embeddings.
\newblock In {\em Proceedings of the 57th Annual Meeting of the Association for
  Computational Linguistics}, pages 371--379, Florence, Italy. Association for
  Computational Linguistics.

\bibitem[Ko and Seo, 2000]{ko-seo-2000}
Ko, Y. and Seo, J. (2000).
\newblock Automatic text categorization by unsupervised learning.
\newblock In {\em Proceedings of the 18th Conference on Computational
  Linguistics - Volume 1}, COLING '00, page 453–459, USA. Association for
  Computational Linguistics.

\bibitem[Lau and Baldwin, 2016]{lau-baldwin-2016-empirical}
Lau, J.~H. and Baldwin, T. (2016).
\newblock An empirical evaluation of doc2vec with practical insights into
  document embedding generation.
\newblock In {\em Proceedings of the 1st Workshop on Representation Learning
  for {NLP}}, pages 78--86, Berlin, Germany. Association for Computational
  Linguistics.

\bibitem[Le and Mikolov, 2014]{Le-Mikolov-2014}
Le, Q. and Mikolov, T. (2014).
\newblock Distributed representations of sentences and documents.
\newblock In Xing, E.~P. and Jebara, T., editors, {\em Proceedings of the 31st
  International Conference on Machine Learning}, volume~32 of {\em Proceedings
  of Machine Learning Research}, pages 1188--1196, Bejing, China. PMLR.

\bibitem[Liu et~al., 2004]{liu2004}
Liu, B., Li, X., Lee, W.~S., and Yu, P.~S. (2004).
\newblock Text classification by labeling words.
\newblock In McGuinness, D.~L. and Ferguson, G., editors, {\em Proceedings of
  the Nineteenth National Conference on Artificial Intelligence, Sixteenth
  Conference on Innovative Applications of Artificial Intelligence, July 25-29,
  2004, San Jose, California, {USA}}, pages 425--430. {AAAI} Press / The {MIT}
  Press.

\bibitem[Mikolov et~al., 2013]{mikolov2013efficient}
Mikolov, T., Chen, K., Corrado, G., and Dean, J. (2013).
\newblock Efficient estimation of word representations in vector space.

\bibitem[Rao et~al., 2006]{RaoPK06}
Rao, D., P, D., and Khemani, D. (2006).
\newblock Corpus based unsupervised labeling of documents.
\newblock In Sutcliffe, G. and Goebel, R., editors, {\em Proceedings of the
  Nineteenth International Florida Artificial Intelligence Research Society
  Conference, Melbourne Beach, Florida, USA, May 11-13, 2006}, pages 321--326.
  {AAAI} Press.

\bibitem[Sievert and Shirley, 2014]{sievert-shirley-2014-ldavis}
Sievert, C. and Shirley, K. (2014).
\newblock {LDA}vis: A method for visualizing and interpreting topics.
\newblock In {\em Proceedings of the Workshop on Interactive Language Learning,
  Visualization, and Interfaces}, pages 63--70, Baltimore, Maryland, USA.
  Association for Computational Linguistics.

\bibitem[Song and Roth, 2014]{Song_Roth_2014}
Song, Y. and Roth, D. (2014).
\newblock On dataless hierarchical text classification.
\newblock In {\em Proceedings of the Twenty-Eighth AAAI Conference on
  Artificial Intelligence}, pages 1579--1585.

\bibitem[Yin et~al., 2019]{yin-etal-2019-benchmarking}
Yin, W., Hay, J., and Roth, D. (2019).
\newblock Benchmarking zero-shot text classification: Datasets, evaluation and
  entailment approach.
\newblock In {\em Proceedings of the 2019 Conference on Empirical Methods in
  Natural Language Processing and the 9th International Joint Conference on
  Natural Language Processing (EMNLP-IJCNLP)}, pages 3914--3923, Hong Kong,
  China. Association for Computational Linguistics.

\bibitem[Zhang et~al., 2015]{zhang-et-al-2015}
Zhang, X., Zhao, J., and LeCun, Y. (2015).
\newblock Character-level convolutional networks for text classification.
\newblock In {\em Proceedings of the 28th International Conference on Neural
  Information Processing Systems - Volume 1}, NIPS'15, page 649–657,
  Cambridge, MA, USA. MIT Press.

\bibitem[Zhang et~al., 2020]{Zhang2020}
Zhang, Y., Meng, Y., Huang, J., Xu, F., Wang, X., and Han, J. (2020).
\newblock Minimally supervised categorization of text with metadata.
\newblock In {\em Proceedings of the 43rd International ACM SIGIR Conference on
  Research and Development in Information Retrieval}, pages 1231--1240.

\end{thebibliography}

\end{document}